\documentclass{article}
\usepackage{spconf,amsmath,graphicx}
\usepackage{url}
\usepackage{float}

\title{SUB-IMAGE RECAPTURE FOR MULTI-VIEW 3D RECONSTRUCTION}
\name{Yanwei Wang\thanks{Thanks to Microsoft Corporation for funding of this work.}}
\address{Mixed Reality\\
     Microsoft Corporation}
\begin{document}
%
\maketitle
\begin{abstract}
3D reconstruction of high-resolution target remains a challenge 
task due to the large memory required from the large input image size. Recently developed
learning based algorithms provide promising
reconstruction performance than traditional ones, however,
they generally require more memory than the traditional algorithms and facing
scalability issue. In this paper, we developed a generic approach, sub-image recapture (SIR), 
to split large image into smaller sub-images and process them individually. 
As a result of this framework, the existing 3D reconstruction 
algorithms can be implemented based
on sub-image recapture with
significantly reduced memory and substantially improved scalability.
\end{abstract}
\section{INTRODUCTION}
\label{sec:intro}

3D object reconstruction from imagery has been an active research area for quite long time, 
with tremendous evolutions have been observed.
Two main steps in a 3D reconstruction pipeline include a sparse modeling  
Structure-from-Motion (SfM) step and a dense modeling step
Multi-View Stereo (MVS).
The SfM undergoes a great deal of transformation and 
improvement over the years. Started from the early work~\cite{10.1007/3-540-61123-1_181,341077},
increasingly large-scale reconstruction systems and variations have been developed~\cite{7298949}.
Readers are referred to ~\cite{Schonberger_2016_CVPR} 
for a survey and implementation of the current SfM algorithm. 
The MVS algorithm
has also been advanced from early traditional 
MVS~\cite{sinha2007multi-view, 5226635, bleyer2011patchmatch, 7410463}
to neural net and learning based algorithms including both stereo
~\cite{DBLP:conf/cvpr/FurukawaCSS10, 9897772,
8296737,https://doi.org/10.48550/arxiv.1703.04309, https://doi.org/10.48550/arxiv.1803.08669, 
https://doi.org/10.48550/arxiv.2004.09548, https://doi.org/10.48550/arxiv.1703.06211} 
as well as MVS
~\cite{Ji_2017, https://doi.org/10.48550/arxiv.1708.05375, https://doi.org/10.48550/arxiv.2208.06674,
https://doi.org/10.48550/arxiv.1908.04422,9010001,https://doi.org/10.48550/arxiv.1804.02505,
https://doi.org/10.48550/arxiv.1902.10556,https://doi.org/10.48550/arxiv.1912.06378,
https://doi.org/10.48550/arxiv.1912.08329,https://doi.org/10.48550/arxiv.2007.07714,
https://doi.org/10.48550/arxiv.2003.13017,hui16msgnet} 
with improved accuracy and/or efficiency 
for various datasets. The learning based MVS algorithms
hold the potential solution to hard problems such as variation of 
illumination, lack of texture or non-Lambertian surface, as well as 
occlusion~\cite{https://doi.org/10.48550/arxiv.2012.01411}. 

However, these learning based MVS algorithms is known to have the scalability 
issue due to their high memory consumption. With today's
increasing camera sensor size as well as the super-resolution techniques~\cite{10.1007/978-3-030-11021-5_16}, the memory requirement increases
geometrically. 
For example, a 10K by 10K resolution airborne image for 3D terrain reconstruction will
use 1.2GB memory assuming floating point resolution for all RGB channels. For such 
scenario, even two-view stereo will has minimum memory requirement of 2.4GB, without
considering the extra cost of any algorithm used. 
When the input image size
is too large, down-sampling of the input image has to be performed to meet the memory requirement.

Considerable efforts have been observed in the literature to address this issue
~\cite{https://doi.org/10.48550/arxiv.2003.13017,https://doi.org/10.48550/arxiv.1703.04309,
https://doi.org/10.48550/arxiv.1804.02505,https://doi.org/10.48550/arxiv.1902.10556,
https://doi.org/10.48550/arxiv.1912.06378,https://doi.org/10.48550/arxiv.1912.08329}. All these
works are trying to
improve the performance from the algorithm point-of-view, with experimental maximum
size \textit{between a few hundreds to a couple of thousands} in width/height. 

The framework introduced in this paper address this scalability issue from a data point-of-view,
independent of the algorithms used. 
This solves the existing learning based MVS algorithm's scalability problem once for all, the
existed learning MVS algorithms can be applied to \textit{any large image size} without 
down-sampling.

\section{SIR: AN UNIVERSAL STRATEGY}
\label{sec:strategy}

We develop a sub-image recapture (SIR) approach which can be used to implement existing
3D reconstruction algorithms with reduced memory requirement and improved efficiency.
In fact, it can be used for most 3D image processing algorithm which utilizes camera
model. 

The idea of sub-image recapture starts from splitting the original image into a matrix of smaller images
which is referred to as sub-images. Unlike down-sampling, 
each sub-image has a smaller image 
size than the native image while maintaining the same native spatial resolution 
of the native image such that there is no loss of details.
Each sub-image is then treated as an independent image which is ``recaptured" by a 
corresponding synthesized camera with its own distinct camera parameters. Through
this process, for each target pixel or area of interest, we have more ``focused" sub-images instead
of the original full field of view images to calculate the depth information. 

The following advantages can
be achieved by this approach: 
1) significant memory reduction for each MVS step 
due to the use of sub-images;
2) under the fixed memory limit, in multi-view geometry stage, more images could be used 
for the same target area
hence the improved depth estimation accuracy; 
3) for some applications where there are only partial overlaps between adjacent images,
sub-image approach can enforce only overlapping part of the sub-images enter the MVS step hence
has more efficient use of memory,
4) improved parallel implementation of the MVS
algorithm due to the fact that the original image is split into sub-images 
which can be processed in parallel. 

\section{SYNTHESIZED RECAPTURE CAMERA}
\label{sec:recapturecamera}

\subsection{Definition of Recapture Camera}
Recapture camera is best explained by dividing
a native digital image into a matrix of sub-images.
Each sub-image is associated with a distinct synthesized recapture camera which has synthesized 
intrinsic and extrinsic parameters mapped from a real camera associated with the native digital image. 
In particular, the synthesized intrinsic parameters for a sub-image's camera include a principal 
point which corresponds to the native 
digital image's principal point re-defined relative to the sub-image's coordinate system origin. 
Re-defining the native principal point relative to the sub-image's coordinate system origin  
spatially compensated all the location differences between the sub-image and its native parent, 
without any error involved. The lens distortion of the native image's 
camera will also be carried over to all its sub-images, and hence the undistortion process.

By associating each sub-image with a distinct synthesized recapture camera, 
each sub-image and associated synthesized recapture camera
can be treated as a separate, full frame, digital image as if captured by 
a real camera in its entirety, only with a smaller field of view. 

\begin{figure}[H]

\begin{minipage}[h]{1.0\linewidth}
  \centering
  \centerline{\includegraphics[width=6cm,angle=270,origin=c]{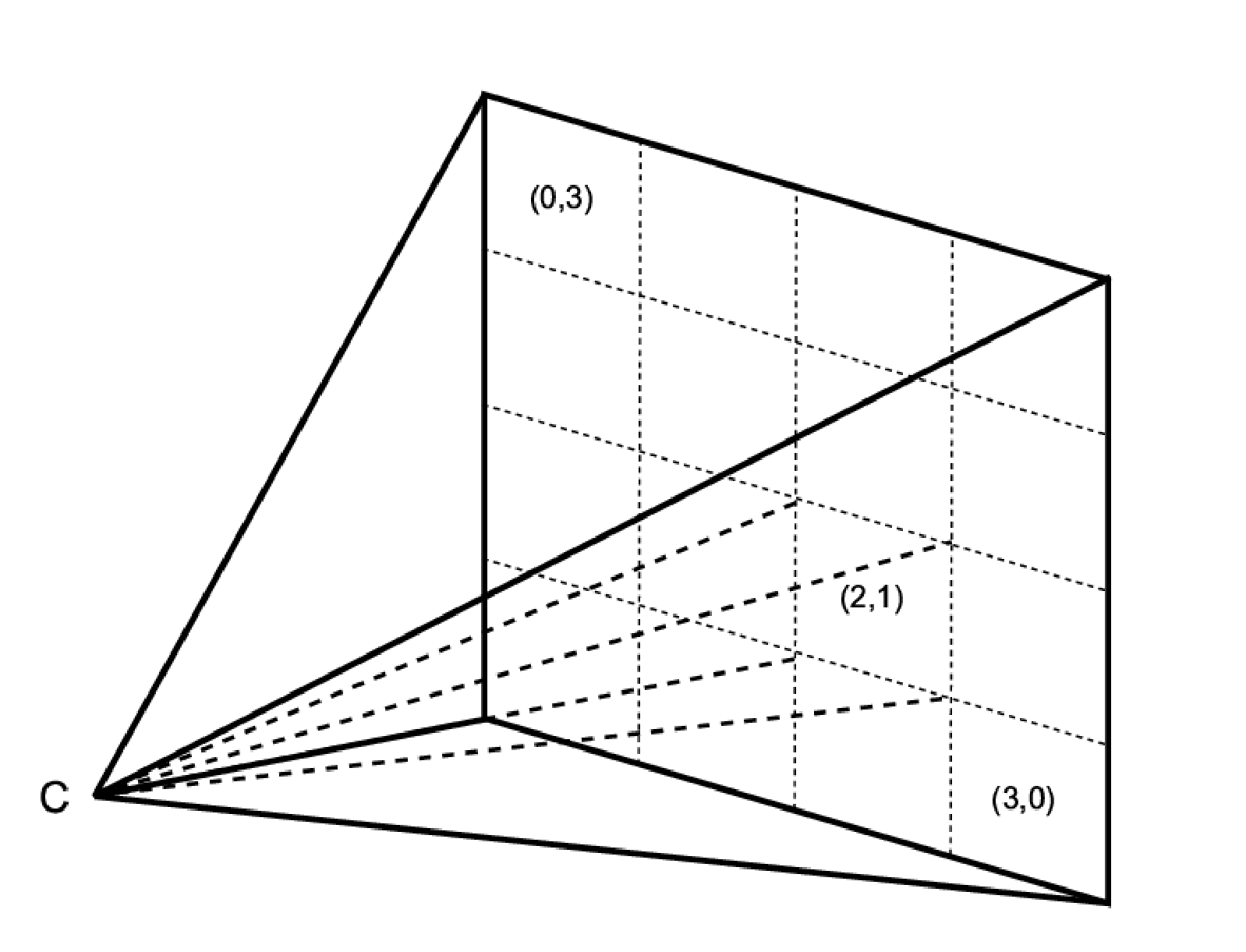}}
  \vspace{-1.0cm}
  \centerline{(a) Camera with 4$\times$4 sub-images}\medskip
\end{minipage}
\begin{minipage}[h]{1.0\linewidth}
  \centering
  \centerline{\includegraphics[width=6cm,angle=270,origin=c]{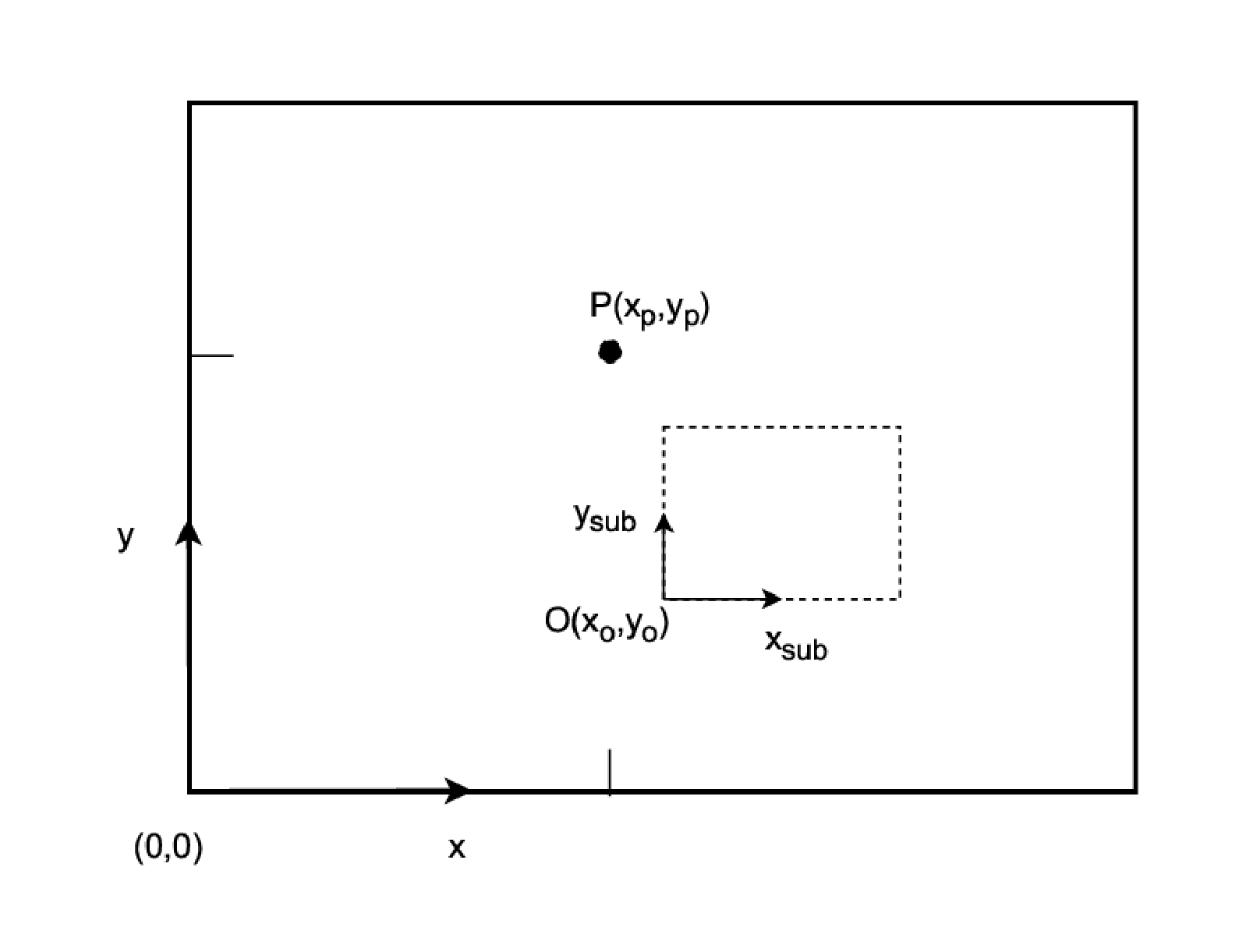}}
  \vspace{-1.0cm}
  \centerline{(b) principal point offset}\medskip
\end{minipage}
\caption{Cameras associated with both the original image and one of 
the corresponding 4$\times$4 sub-images.}
\label{fig:recapture}
\end{figure}

Figure \ref{fig:recapture}(a) illustrates a camera model with 4$\times$4 sub-images. As an example,
the synthesized camera of sub-image index (2,1) is shown in 
Figure \ref{fig:recapture}(a) and \ref{fig:recapture}(b).
Here $O(x_o,y_o)$ denotes the location of the sub-image (2,1)'s origin, $P(x_p,y_p)$ denotes the 
native principal point. The synthesized camera for sub-image (2,1) has its 
principal point defined as 
\begin{equation}
P_{(2,1)}(x,y) = (x_p - x_o,y_p - y_o)
\end{equation}
Note that the synthesized camera principal point could have a negative location value.

So for a native image size of $S_x \times S_y$ with a sub-image recapture configuration of $I \times J$,
the principal points for each sub-image $(i,j)$ is given by
\begin{equation}
P_{(i,j)}(x,y) = (x_p - i \cdot \frac{S_x}{I}, y_p - j \cdot \frac{S_y}{J})
\end{equation}
where $i$ and $j$ denotes the index of sub-image at horizontal and vertical directions, respectively.

The elegance of sub-image recapture is, except for the principal point change, all other intrinsic and 
extrinsic camera parameters of the synthesized recapture camera associated with each
sub-image are kept the same as their corresponding native camera parameters.

Note that depending on application, sub-image recapture may be used to ``recapture" an 
\textit{arbitrary area and size} 
of the native image, not necessarily on a uniform grid of same sized sub-images.

\subsection{SIR for SfM and MVS}
\label{ssec:irforsfmmvs}

The SIR method can be applied to SfM step too. Generally speaking, higher resolution
image will have better feature extraction and matching accuracy performance compared 
with its down sampled 
and blurred counterpart. However, if there are a lot of overlaps and less challenges between clustered 
images, the down sampled image may still work for SfM, thanks to the scale-invariant
feature extraction and the powerful
state-of-the-art SfM algorithms. Another observation from SfM is, it usually requires a single
image each time during the feature extraction step, hence 
in terms of memory usage it is not as demanding as MVS, which requires
a group of images work together to infer the spatial information.

Since for the images we are experimenting with, the majority of the SIR improvements comes
from MVS step, in this paper, we will focus on its performance on dense modeling.



\subsection{SIR FOR 3D RECONSTRUCTION}
\label{ssec:reconstruction}

Here we present a typical complete 3D pipeline for high-resolution reconstruction
with sub-image recapture.

\begin{figure}[ht]

\begin{minipage}[h]{1.0\linewidth}
  \centering
  \centerline{\includegraphics[width=6.5cm,angle=270,origin=c]{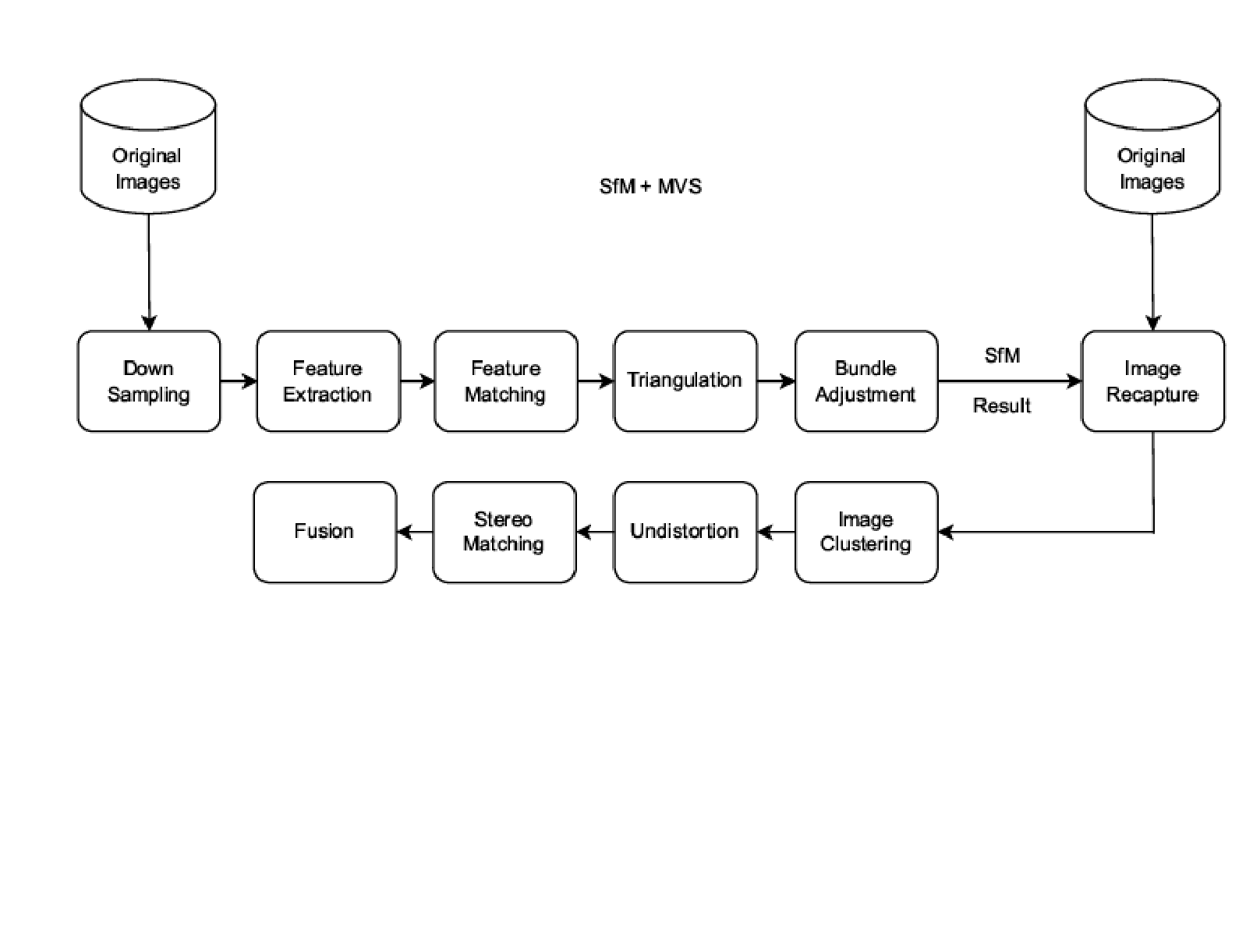}}
  \vspace{-3.0cm}
\end{minipage}
\caption{3D reconstruction pipeline.}
\label{fig:pipeline}
\end{figure}

Figure \ref{fig:pipeline} shows the reconstruction pipeline used in the experiment
with SfM followed by MVS step. 
Due to the large image size, the input to the SfM is decimated 
to a maximum size of 2304 to render a manageable
image size for SfM. For the reason discussed in Section ~\ref{ssec:irforsfmmvs}, we did not 
apply sub-image recapture for SfM in this dataset.

Some details of the pipeline: an incremental mapping~\cite{Schonberger_2016_CVPR}  is used in SfM;
an image clustering~\cite{5235143} step is used to group the overlapping 
sub-images for MVS; the recently proposed 
PatchmatchNet~\cite{https://doi.org/10.48550/arxiv.2012.01411} has been used as the Stereo Matching
step. 
For more details of each block in the pipeline, please refer to 
COLMAP~\cite{Schonberger_2016_CVPR}, PatchmatchNet, and 
their related references.

\section{EXPERIMENTS}
\label{sec:experiments}

\subsection{Datasets}

The data we used here is from a high altitude airborne terrain data provided by 
Vexcel Imaging, which covers the area of Angel Island, CA. The camera rig 
contains 5 cameras facing forward, backward, left, right, and nadir. The sizes of
these native images are 10300$\times$7700, 
7700$\times$10300, and 13470$\times$8670, with an average
resolution of 7cm.

\subsection{Implementation Details}

Our implementation is based on the well-established open-source 3D reconstruction 
software COLMAP. A fork of COLMAP has been developed
to implement sub-image recapture. 
PatchmatchNet software is used as the stereo matching step.

The Angel Island has been divided into small tiles according to the Bing map tile system. 
Each tile at Level of Detail 
(LOD) 19 is processed with an NVIDIA GeForce RTX 2080ti with 11GB memory. 
At LOD 19, each tile has a size of approximately 40x40 meters. 

A group of down-samplings between 5$\times$ to 6$\times$ are applied to the input images of SfM. 
The image clustering step is configured to generate a cluster size around 20.

The sub-image recapture is configured to have 5$\times$5 sub-images for each native image. 
This configuration is chosen to have 2.3K maximum
image size, which is about the upper limit our current MVS pipeline can handle. 
To compare the MVS result of 
SIR vs. without SIR (decimation),
a down-sampling (resize) step is used to makes sure that the 
MVS input image size has a maximum
of 2304 pixels. 

\subsection{Experimental Results}

\begin{figure}[H]

\begin{minipage}[h]{.48\linewidth}
  \centering
  \centerline{\includegraphics[width=4.0cm]{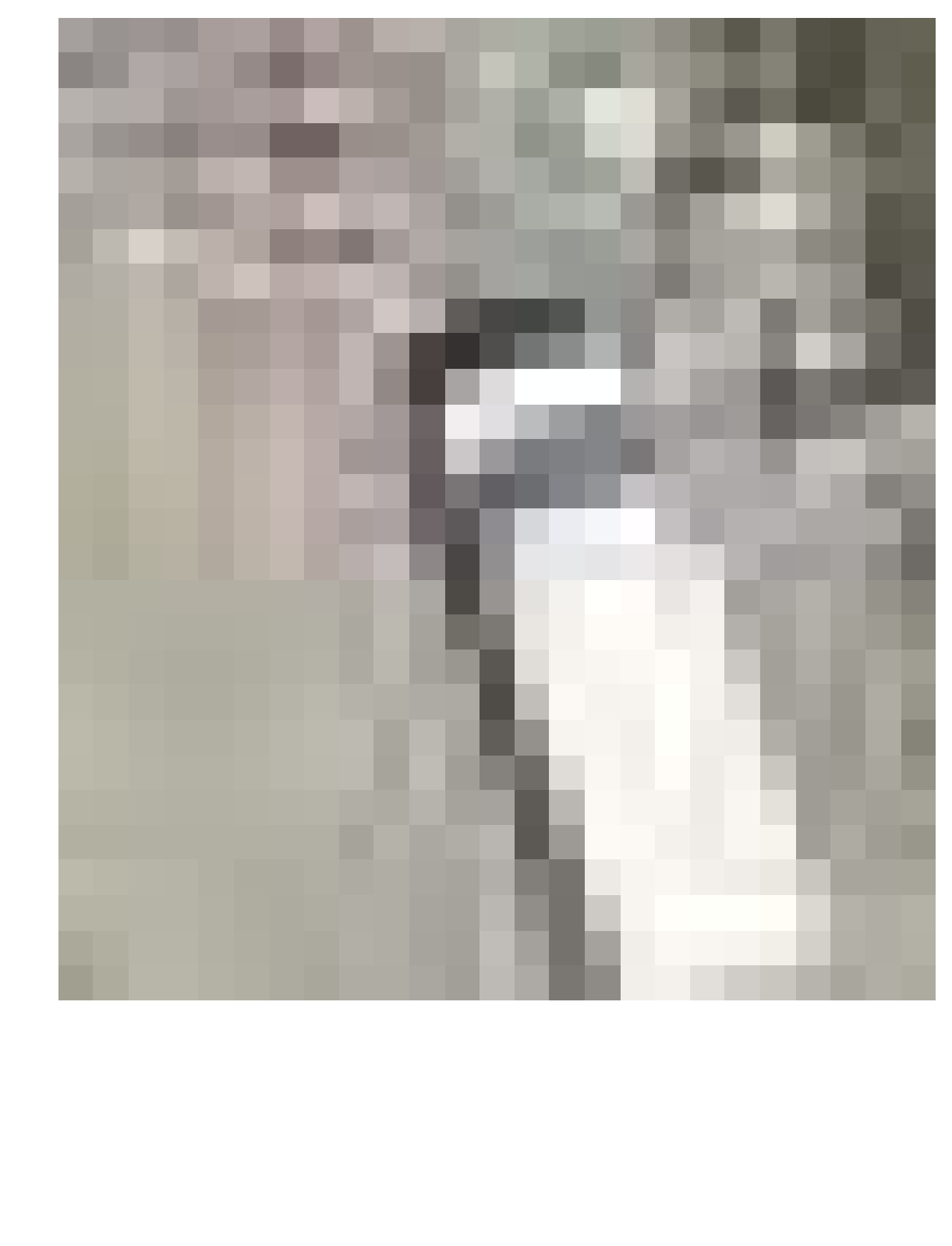}}
  \vspace{-0.5cm}
  \centerline{(a) Down-sampling}\medskip
\end{minipage}
\begin{minipage}[h]{0.48\linewidth}
  \centering
  \centerline{\includegraphics[width=4.0cm]{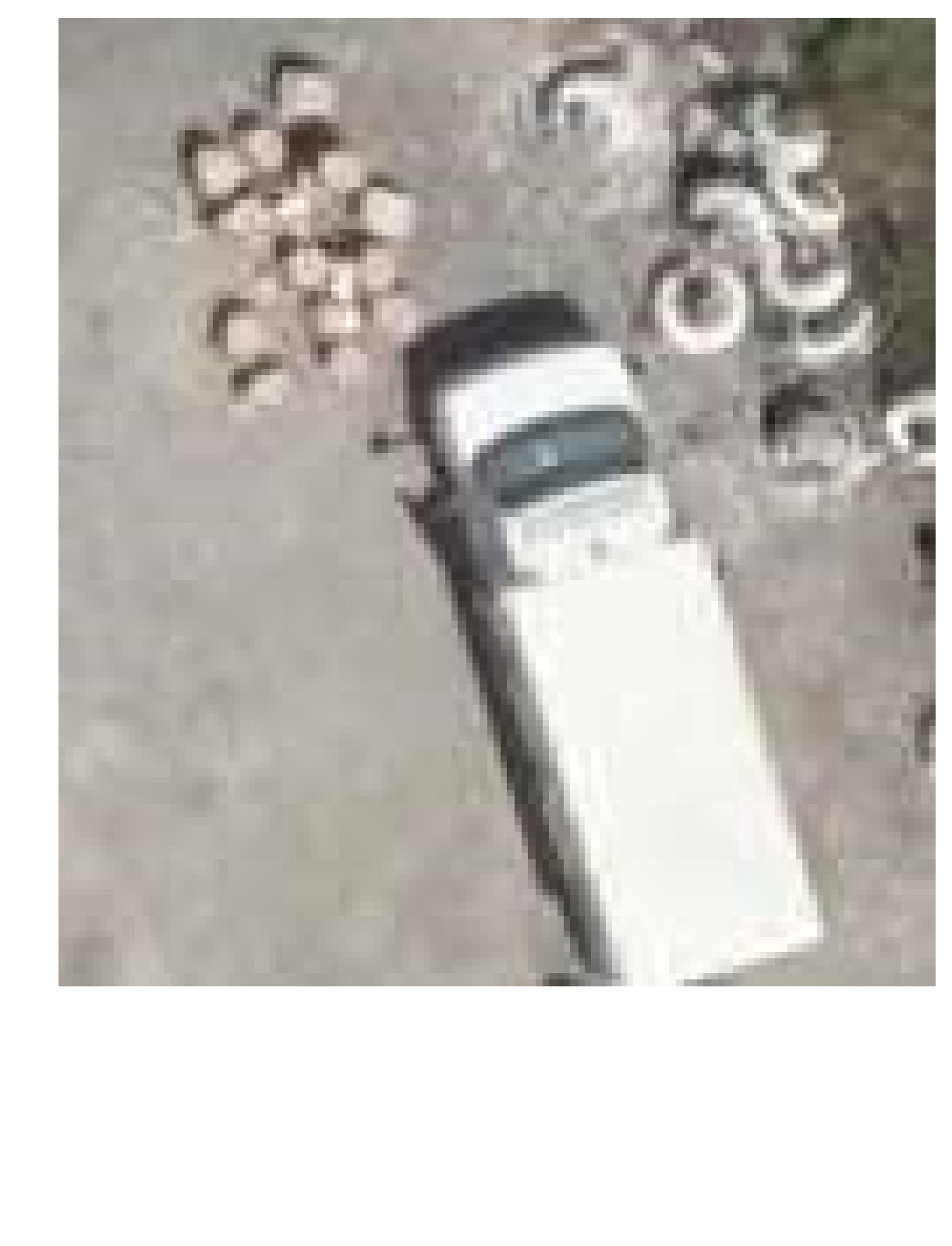}}
  \vspace{-0.5cm}
  \centerline{(b) Native}\medskip
\end{minipage}
\begin{minipage}[h]{1.0\linewidth}
  \centering
  \centerline{\includegraphics[width=7cm]{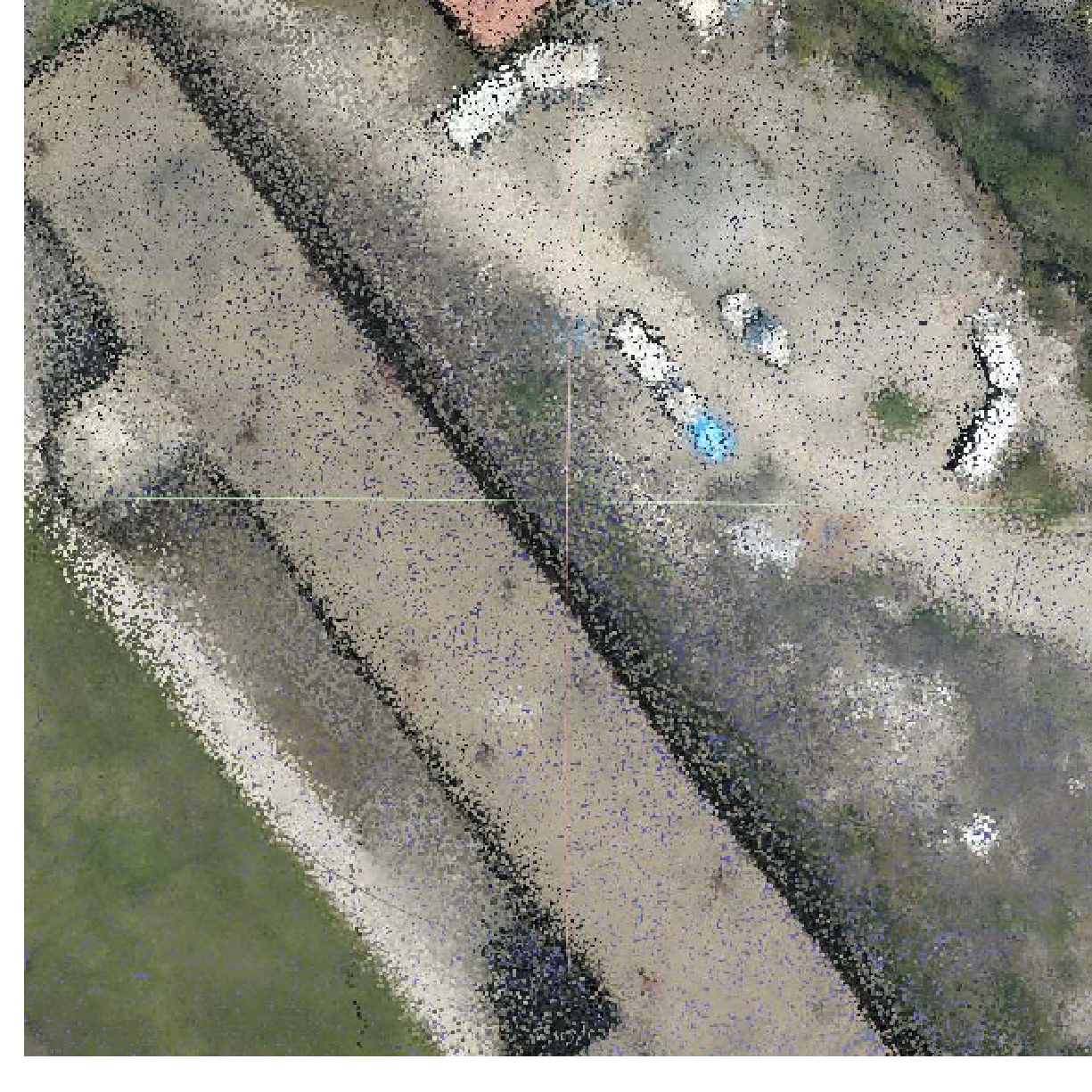}}
  \centerline{(c) without SIR}\medskip
\end{minipage}
\begin{minipage}[h]{1.0\linewidth}
  \centering
  \centerline{\includegraphics[width=7cm]{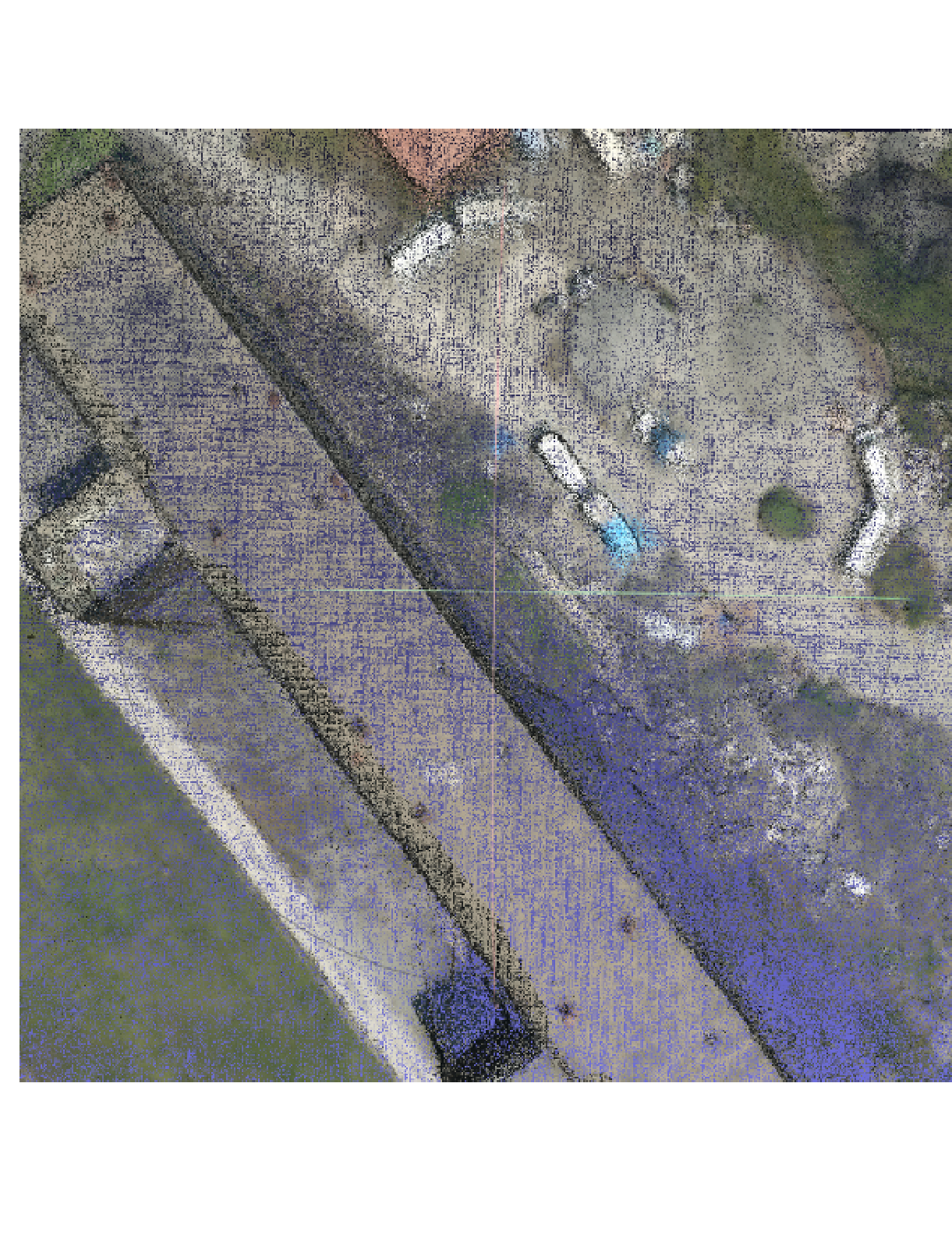}}
  \vspace{-0.7cm}
  \centerline{(d) with SIR}\medskip
\end{minipage}
\caption{A tile of the 3D terrain reconstruction point cloud.}
\label{fig:resimage}
\end{figure}

Figure \ref{fig:resimage}(a) and (b) show the comparison of a highly zoomed-in area of a 
target tile: decimated image with a close to a 5$\times$ stride vs. SIR with native resolution. 
Clearly, the decimated image shows blurred target features.

Figure \ref{fig:resimage}(c) and (d) compare the reconstruction result point cloud between
without SIR and with SIR, respectively, for a tile. 
SIR demonstrates
much better reconstruction results due to the high-resolution sub-images it utilized.

\begin{figure}[H]

\begin{minipage}[H]{1.0\linewidth}
  \centering
  \centerline{\includegraphics[width=8cm]{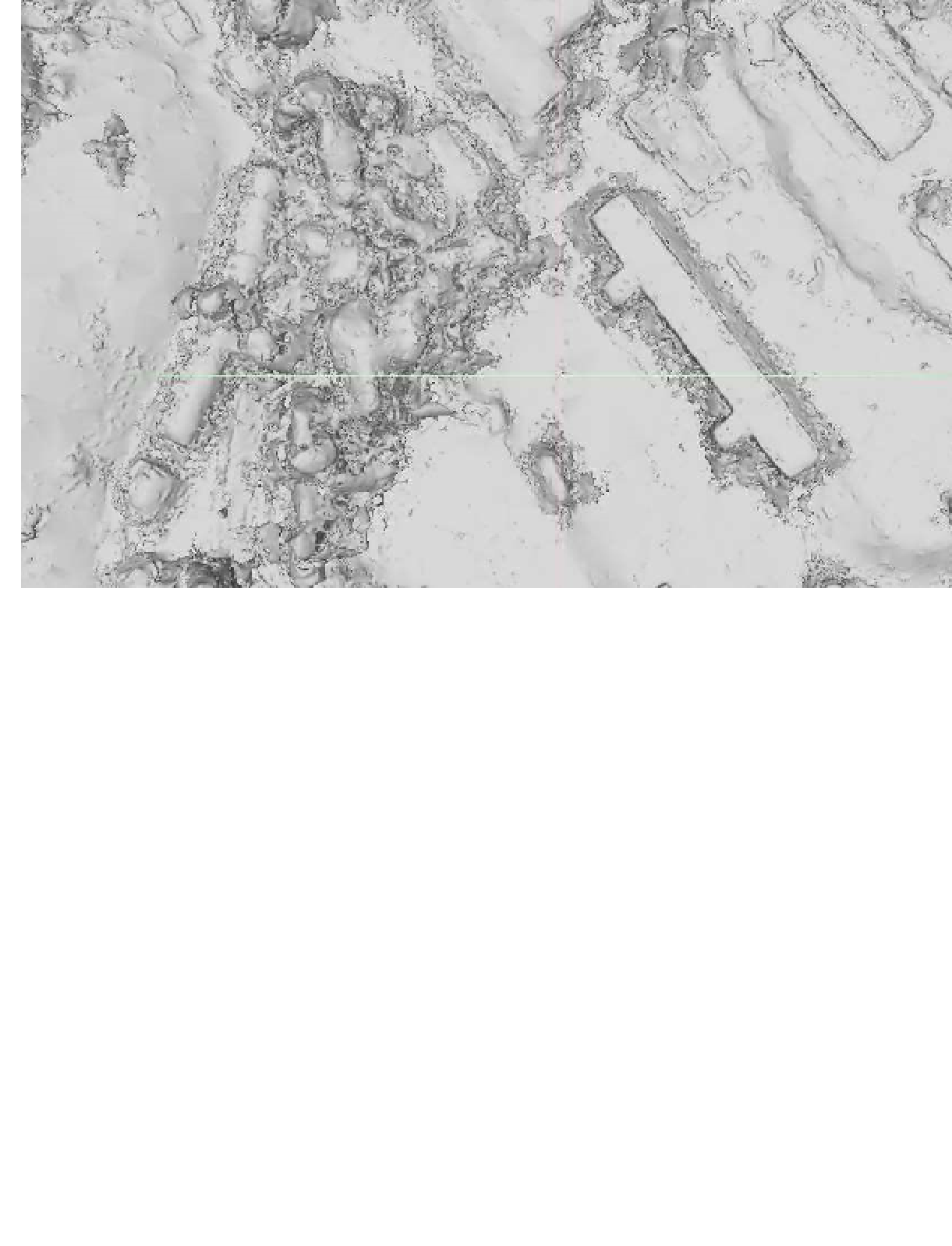}}
  \vspace{-5.0cm}
  \centerline{(a) Down-sampling}\medskip
\end{minipage}
\begin{minipage}[H]{1.0\linewidth}
  \centering
  \centerline{\includegraphics[width=8cm]{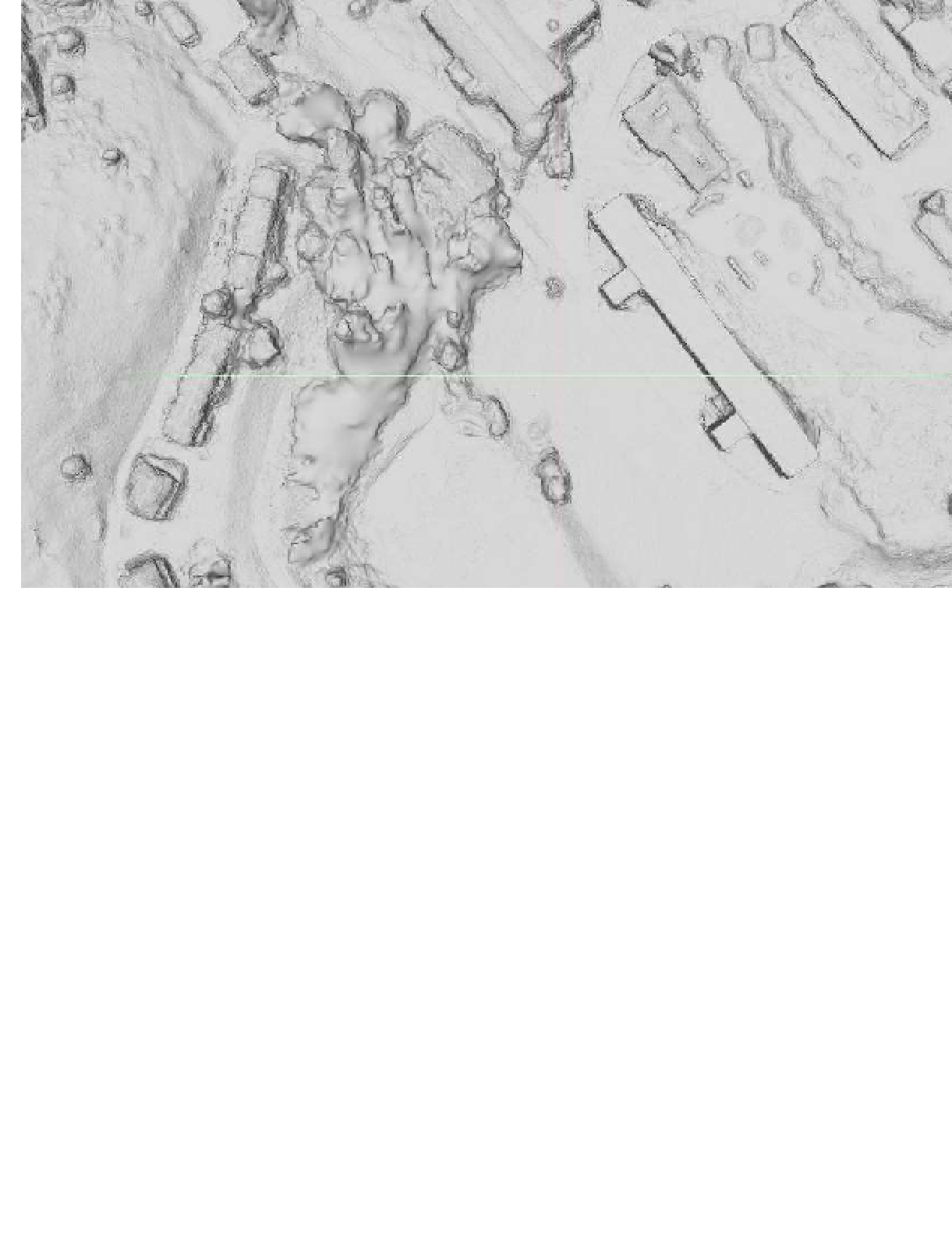}}
  \vspace{-5.0cm}
  \centerline{(b) Sub-Image Recapture}\medskip
\end{minipage}
\caption{Mesh of a small area of the reconstructed 3D terrain.}
\label{fig:resmesh}
\end{figure}

Figure \ref{fig:resmesh}(a) and (b) show the Poisson 
surface reconstructed meshes from the result point cloud,
without SIR and with SIR, respectively, by combining
the output from multiple reconstructed tiles. The area shown is located at the east side of the island with
buildings, vehicles, and trees. 
SIR's mesh shows much clean result with better accuracy and resolution.

\section{CONCLUSIONS}
\label{sec:conclusions}

A Sub-Image Recapture (SIR) framework has been proposed and applied to 
high level-of-detail 3D reconstruction applications. Compared with the conventional
down-sampling of the input image, SIR significantly outperforms its counterpart 
by retaining the native image's resolution without degrading the image quality. 
To the best of our knowledge, this is the first time that state-of-the-art learning 
based MVS algorithm has been applied to the large images 
(13470$\times$8670) used in the paper. 
Fundamentally, SIR solved the current memory-limited scalability problem and enabled the 
MVS algorithms to be applied to \textit{``arbitrarily large"} images. 
It also provided improved accuracy and implementation efficiency.
Based on COLMAP and PatchmatchNet, 
SIR is implemented as an open-source software and released to the public.

\nopagebreak[4]

\bibliographystyle{IEEEbib}
\bibliography{strings,refs}

\end{document}